\documentclass[conference]{IEEEtran}
\IEEEoverridecommandlockouts
\usepackage{cite}
\usepackage{amsmath,amssymb,amsfonts}
\usepackage{algorithmic}
\usepackage{graphicx}
\usepackage{textcomp}
\usepackage{xcolor}
\def\BibTeX{{\rm B\kern-.05em{\sc i\kern-.025em b}\kern-.08em
    T\kern-.1667em\lower.7ex\hbox{E}\kern-.125emX}}
\usepackage[utf8]{inputenc}
\usepackage{array}
\usepackage{url}
\usepackage{listings}
\usepackage{subcaption}
\usepackage{verbatim}
\usepackage{times}
\usepackage{multicol}
\usepackage{multirow}
\usepackage{ntheorem}
\newtheorem{research-question}{Research Question}

\title{Sound Judgment: Properties of Consequential Sounds Affecting Human-Perception of Robots}

\author{\IEEEauthorblockN{Aimee Allen}
\IEEEauthorblockA{\textit{Faculty of Engineering}\\
\textit{Monash University}\\
Melbourne, Australia\\
aimee.allen@monash.edu}
\and
\IEEEauthorblockN{Tom Drummond}
\IEEEauthorblockA{\textit{School of Computing and Information Systems}\\
\textit{University of Melbourne}\\
Melbourne, Australia\\
tom.drummond@unimelb.edu.au}
\and
\IEEEauthorblockN{Dana Kuli{\'c}}
\IEEEauthorblockA{\textit{Faculty of Engineering}\\
\textit{Monash University}\\
Melbourne, Australia\\
dana.kulic@monash.edu}
\thanks{This research was partly supported by an Australian Government Research Training Program Scholarship. D. Kuli{\'c} is supported by the ARC Future Fellowship (FT200100761).}
}

\begin{document}

\maketitle

\begin{abstract}
Positive human-perception of robots is critical to achieving sustained use of robots in shared environments. One key factor affecting human-perception of robots are their sounds, especially the consequential sounds which robots (as machines) must produce as they operate. This paper explores qualitative responses from 182 participants to gain insight into human-perception of robot consequential sounds. Participants viewed videos of different robots performing their typical movements, and responded to an online survey regarding their perceptions of robots and the sounds they produce. Topic analysis was used to identify common properties of robot consequential sounds that participants expressed liking, disliking, wanting or wanting to avoid being produced by robots. Alongside expected reports of disliking high pitched and loud sounds, many participants preferred informative and audible sounds (over no sound) to provide predictability of purpose and trajectory of the robot. Rhythmic sounds were preferred over acute or continuous sounds, and many participants wanted more natural sounds (such as wind or cat purrs) in-place of machine-like noise. The results presented in this paper support future research on methods to improve consequential sounds produced by robots by highlighting features of sounds that cause negative perceptions, and providing insights into sound profile changes for improvement of human-perception of robots, thus enhancing human robot interaction.
\end{abstract}

\begin{IEEEkeywords}
    robot consequential sounds; psychoacoustics; human-centered robotics; social HRI
\end{IEEEkeywords}

\section{Introduction}
As robots increasingly appear in more human occupied environments such as public spaces, workplaces and homes, having positive human-perception of these robots is critical to achieve sustained use of robots. Within these shared environments, robots and humans are typically in close proximity, and so how people think and feel about robots matters to successful human robot interactions (HRI) and adoption. As sound is a key element for human communication, ability to concentrate and mood~\cite{Jariwala2017,DePaivaVianna2015,Jouaiti2019,Basner2014}, sounds made by robots are a key factor affecting human perception of robots~\cite{Allen2024e1p1}. During normal operation, robots produce \emph{consequential sounds}, with many current robots being quite loud or producing unpleasant sounds, which can negatively impact human robot interactions.

\begin{figure}[t]
    \centering
    \includegraphics[width=0.8\linewidth]{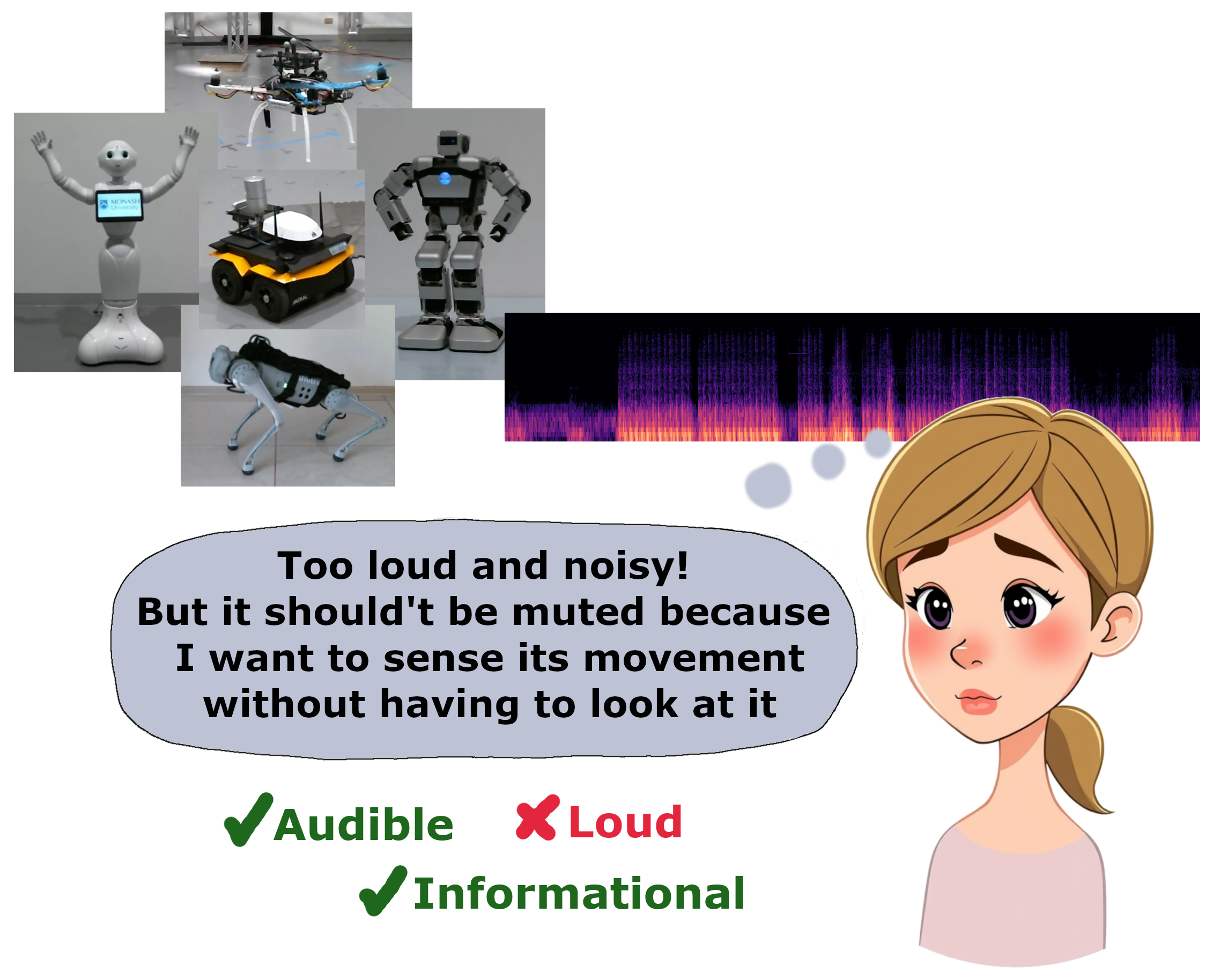}\hfill
    \caption{What properties of robot consequential sounds affect human-perceptions of robots, either negatively or positively?}
    \label{fig:visual_paper_summary}
\end{figure}

Consequential sounds are `sounds that are generated by the operating of the product itself'~\cite{Langeveld2013} meaning the `unintentional noises that a machine produces as it moves and operates'~\cite{Allen2023}. For robots, these are primarily the audible `noises' made by the robot's actuators (such as sound from an electric motor moving a robot joint), meaning the robot can not function without generating consequential sounds. The types of consequential sounds produced by a robot vary depending on elements such as robot form factor, type of actuator and how the robot is moving\cite{Wang2016,Allen2023}. Similar machine sounds from air-conditioners and vehicles in human environments are known to cause distraction, annoyance and reduced productivity~\cite{Basner2014,DePaivaVianna2015,Jariwala2017}.

With robots entering shared human spaces, it is important to understand how people perceive robot consequential sounds, why negative perceptions exist, and hence options to improve them. Whilst consequential sounds cannot be prevented, choices can be made to alter how people perceive them, such as selecting robot components which generate more pleasant sounds, utilising dampening or noise cancelling techniques, or adding sound augmentations to disguise unwanted attributes.

\noindent This paper makes the following contributions:
\begin{itemize}
    \item An exploratory analysis of qualitative responses from a large, diverse cohort, identifying key properties of consequential sounds correlating with negative and positive human-perceptions.
    \item Insights into what sounds people want to hear (and not hear) from robots, informing potential methods to augment robot consequential sounds to improve human perceptions of robots.
\end{itemize}

\section{Related Work} \label{sec:related-works}
Consequential sounds exist for every robot and thus every human-robot interaction, however the study of consequential sound is still rare within HRI research. A recent review of sound in robotics identified only seven consequential sounds papers\cite{Zhang2023}. Psychoacoustics research (the study of how humans perceive sound) has shown that sound is a critically important sensory element for humans~\cite{Jouaiti2019}, with the ability to contribute both positive (happiness, relaxation) or negative effects (anxiety, distraction, annoyance). Human response to different sound intensities and frequencies varies, with loud noises and high-pitched sounds being disliked~\cite{Cha2018}. Broadband noises are generally perceived more pleasantly than pure tones~\cite{Fastl2007a} particularly those with many frequencies evenly spaced across the human-audible range~\cite{Cha2018}. Humans are very familiar with sharing environments with biological agents such as people or animals, and are thus accustomed to regular, rhythmic sounds such as breathing or rustling denoting an embodied agent's presence via proxemics~\cite{Jouaiti2019,Trovato2018}. In contrast, acute sounds such as alarms, or the sudden startup noises of a machine tend to be disliked, often eliciting an instinctive danger response~\cite{Jouaiti2019}. Sound perception is often entwined with other sensory perception such as visual perception. There are often bi-directional links between audio and visual stimulus\cite{DeGelder2000}, with different sound profiles (e.g. rhythmic versus acute sounds) creating a crossmodal effect from audio to visual perception\cite{Vroomen2000}.

Sounds have been shown to have an impact on human robot interactions by influencing opinions of particular robots~\cite{Moore2018,Frederiksen2019,Frederiksen2020,Cha2018,Song2017}, causing confusion or annoyance~\cite{DePaivaVianna2015,Schute2007,Moore2017} or positively contributing a sense of proxemics~\cite{Cha2018,Trovato2018}. The earliest paper to focus on robot ``consequential sound''~\cite{Moore2017}, compared videos of pairs of DC motors with artificially dubbed sounds to gauge participant preferred sounds, showing consistency within-participants but not between-participants. Several studies have examined how the perception of robotic arms varied when their consequential sounds were altered~\cite{Zhang2021,Tennent2017}. In one experiment, frequency (pitch) and intensity (volume) of consequential sounds were manipulated~\cite{Zhang2021} with results suggesting that quieter robots are preferred, and increased sound frequency correlated with perceptions of warmth. Another experiment dubbed consequential sounds from a low quality robotic arm onto videos a KUKA robotic arm, finding that both sets of consequential sounds reduced robot quality ratings, but with the higher quality robot sounds preferred~\cite{Tennent2017}.

Consequential sounds have also been shown to interact with other sounds and robot motions to confuse the interpretation of affect. For example, low-frequency consequential sounds created a negative valence and strong arousal affect such that the robot was perceived as frustrated and overshadowed the intended affect~\cite{Frid2018,Frid2022}. In contrast, another study used headphones to dampen existing consequential sounds and play additional affective sounds, leading to the robot being perceived as curious, happy and less angry~\cite{Frederiksen2019}. Recent research projects have investigated improving opinions of robots by adding transformative sounds on-top of the consequential sounds~\cite{Zhang2021b,Robinson2021,Wang2023}. One study measured perception of competence and warmth of a robot, using a within-participant study comparing a robot's consequential sounds with and without transformations~\cite{Zhang2021b}. In a second experiment, carefully designed sounds were added to consequential sounds, and compared to a silent control~\cite{Robinson2021}, however comparison to unedited consequential sounds was not made. Another experiment added natural sounds to the consequential sounds of a micro-drone~\cite{Wang2023} leading to a more pleasant perception of the drone.

The effect of consequential sounds on human-centric perception of robots (rather than perception of the robot itself) was examined in a recent online study~\cite{Allen2024e1p1}. Using a between-participant design, participants were shown videos of robots either with or without sound, with the presence of consequential sounds being shown to significantly increase negative human-perception of robots, including feeling more distracted, increasing negative associated affects, and being less willing to colocate with robots. Another recent work provides a detailed explanation of consequential sounds within the context of robotics, including issues with how people may perceive these sounds, and how to manage consequential sounds~\cite{Allen2023}.

Existing research has helped to illustrate the effect that consequential sounds can have within human-robot interactions, however very few studies have large samples, almost all focus on robot-centric attributes (such as trusting the robot) and lack insight into the properties of the sounds causing these perceptions. This paper builds on existing research by contributing insights into the properties of consequential sounds influencing the phenomenon of human-centric perceptions of robots, thus providing a strong starting point for augmenting and improving these sounds, and by extension improving human-perception of robots and HRI.

\section{Research Methods}
\subsection{Research Questions} \label{sec:research-questions}
Building on prior work~\cite{Allen2024e1p1} indicating that consequential sounds create negative human perceptions of robots, this paper examines what properties of the sound cause these perceptions, and what human observers indicate can be changed to improve perceptions. We explore the following specific research questions:

\begin{research-question}(RQ1) \label{research-question1}
    What properties of robot consequential sounds affect human-perceptions of robots, either negatively or positively?
\end{research-question}

\begin{research-question}(RQ2) \label{research-question2}
    How would people prefer robots to sound instead?
\end{research-question}

\subsection{Experiment Overview}
To answer the above research questions, an on-line survey was conducted attracting a large cohort of diverse participants. The experiment was comprised of: a pre-questionnaire of demographic and robot-familiarity questions, sound tests to verify participants' ability to hear sounds, 5 trials each consisting of a 20 second video of a robot performing its typical movements, per-trial questions, and a final post questionnaire. The online Qualtrics platform was used to run the experiment with participants averaging 30 minutes to complete. Running the experiment online provided for a larger and more diverse cohort, and also facilitated having a `no sound' control condition without changing other experiment parameters such as having no motion on the robot.

Participants were randomly assigned (50/50 condition split) between the consequential sound (CS) group who saw 5 videos of robots with sound, and the control group of no-sound (NS) who saw identical videos but with the sound removed. Providing these two conditions allowed insights to be gathered on preferred sounds from sound-group participants (who may have been biased by the sounds in the existing videos), and the no-sound participants without this potential bias. The no sound condition is also useful for comparison to related work~\cite{Robinson2021, Izui2020} which often has a silent condition. The study protocol was approved by Monash University Human Research Ethics Committee (MUHREC) (Project ID 36414).

The robots used as the stimuli were selected to a) be robots that are likely to exist within human-occupied environments (thus producing sound stimulus in human vicinity), b) be diverse in terms of form factors, uses, motions and sounds produced, and c) include few enough robots/trials to avoid participant fatigue. Fig.~\ref{fig:visual_paper_summary} includes images of the 5 robots used in this experiment; a custom Quadrotor drone, Pepper social humanoid (SoftBank Robotics), Jackal mobile UGV (ClearPath Robotics), Yanshee table-top humanoid (UBTECH), and Go1 EDU PLUS quadruped (Unitree). For this experiment, the Pepper and Yanshee robots utilised pre-programmed movement routines, and the other 3 robots were teleoperated to perform the desired movements (and consequential sounds) for the experiment. Quiet filming environments were selected so the robot sound was dominant, with the background noise (spectra and intensity) typical of an office or home environment. It is important for the trial recordings to replicate a real-world scenario by maintaining normal sound context for the background sound, as environment sounds may alter perception by either amplifying responses or acting as a mask for the robot consequential sounds. Participants were presented with contextual information to match common attributes of many HRI deployments, both implicitly in terms of where and how videos were filmed, and explicitly within the questionnaires e.g. specifying collate with robot (but not direct interaction), and within home, workplace, or public-space.

The qualitative analysis presented in this paper utilises the same experimental data as Allen et al.\cite{Allen2024e1p1} where the data was quantitatively analysed.

\subsection{Qualitative Questions} \label{sec:questions-and-priming}
Participants were asked to comment on their perceptions of the robots after seeing each robot video (4 questions) and in an overall post-questionnaire (7 questions). Participants were not informed that the premise of this experiment was to study sound until the very end of the experiment (during the post questionnaire). This avoided participants consciously focusing on the sounds, and facilitated accounting for multi-modal perception effects between audio-visual stimuli. To disguise this sound premise, participants were asked additional non-sound questions including asking about different robot attributes (visual, movement, audible) and including sound tests to confirm ability to hear without revealing the purpose of the study. This was done to facilitate examining the differences in feedback spontaneously provided on perceptions towards sound (prior to reveal) versus when asked specifically for their opinions on the sounds. The questions were thus separated into two groups based on if the participant knew they were answering about sound, i.e. 'Sound Primed Questions' versus general robot questions. Short forms of the questions can be seen in Table~\ref{tab:questions}, with full questionnaires on the research project website\footnote{Supplementary material, robot videos and questionnaires available from \url{https://aimeeallen.github.io/perception-of-robot-consequential-sounds/}}.

\begin{table}
    \centering
    \caption{`General' and `Sound Primed' questions appeared either Per trial or during the Post questionnaire (Post-Q). `CS ONLY' were presented to sound condition participants only.}
    \label{tab:questions}
    \begin{tabular}{|p{0.13\linewidth}|p{0.46\linewidth}|p{0.21\linewidth}|}
	\hline
	\textbf{Question Priming}&\textbf{Question Topic}&\textbf{Source}\\\hline
        \multirow{5}{0.13\linewidth}{General Questions} & Anything specific that you liked or disliked about the robot? & Per trial (Q10.3)\\\cline{2-3}
	& What features of this robot make you happy or unhappy to be in the same space as this robot? & Per trial (Q12.2)\\\cline{2-3}
        & Other comments (about robot) & Per trial (Q12.3)\\\cline{2-3}
        & Why favourite robot & Post-Q (Q50.1)\\\cline{2-3}
        & Why least favourite robot & Post-Q (Q50.2)\\\hline
        \multirow{6}{0.13\linewidth}{Sound Primed Questions} & Describe how you would want this robot to sound & Per trial (Q11.3)\\\cline{2-3}
	& (CS ONLY) Why favourite robot sounds & Post-Q (Q52.1)\\\cline{2-3}
        & (CS ONLY) Why least favourite robot sounds & Post-Q (Q52.2)\\\cline{2-3}
        & (CS ONLY) Overall feelings on consequential sounds & Post-Q (Q52.3)\\\cline{2-3}
        & Sounds want robots to make & Post-Q (Q53.2)\\\cline{2-3}
        & Any other comments (final experiment question) & Post-Q (Q53.3)\\\hline
    \end{tabular}
\end{table}

\subsection{Analysis methodology}
Qualitative data was analysed using the NVivo qualitative data analysis (QDA) tool. Category analysis techniques\cite{Linneberg2019,Kiger2020} were used to perform topic modelling in a blended method approach. Topic modelling was chosen as quantified counts of topic codes provide the best exploratory insights on how to meet the shared needs of a broad population~\cite{Vaismoradi2013}, thus providing a good baseline for building further individualised customisations.

The analysis involved several iterative coding steps across all questions. Firstly, the data was examined inductively for prevalent codes. This included thematic auto coding of all the data, word cloud analysis, and manual reading of large samples to uncover pilot codes missed by the auto coding. The inductive codes generated were combined with domain knowledge and filtered for topics relevant to research questions, leading to the creation of 16 sound property topic codes of interest and 4 valence codes (see results section~\ref{sec:topic-analysis}). In the second main iteration, these deductive topic codes were used to manually code 35\% of the data (65 participants/300 trials). NVivo's `pattern match' auto-coding tool was then used to code the remaining data at a per-sentence level. Within this tool, `code more' was initially used, but then restricting the `terms used to find new coding references' to only include terms that a human coder would decide were relevant for that code. Auto-coded results were then manually reviewed for any missed or misaligned results leading to 16 incorrect codes being re-adjusted.

In order to explore Research Question\ref{research-question1} (RQ1) on liking and disliking existing consequential sounds, additional analysis was performed on the spectral features of the experiment sound stimuli robot sounds (see section~\ref{sec:linking-to-spectrogram}). Common coded sound properties and valences from sound-condition participants were compared with the observed sound features to see what spectral features align with the identified topics, and what spectral features are not mentioned (or infrequently mentioned) and thus did not have a noted effect of perception. To address Research Question\ref{research-question2} (RQ2), responses from all participants (both sound and control conditions) were examined to gain insights as to what sounds people want, and what to avoid, thus providing insights into how consequential sounds can be altered for future robot deployments (e.g. cancelled, augmented, masked, components chosen).

\section{Results}
\subsection{Participant Diversity}
Participant recruitment began with global networks of researchers (e.g. industry partnerships, social media), with snowballing across all channels, and less than 3\% from survey-share platforms (incentivised). A total of 186 participants were recruited, completing 878 trials. Results of the sound calibration test were used to verify participant's ability to hear the sound stimuli. Data was removed for the four participants (2\%) who failed the sound test (2 or more questions incorrect), resulting in 858 trials across 182 participants. Of these participants, 89 (48.9\%) experienced the sound condition, with the remaining in the control condition (completely silent videos of robots). Good diversity was achieved within this sample population across age groups: 18 - 24 (13.7\%), 25 - 34 (19.8\%), 35 - 44 (19.8\%), 45 - 54 (12.1\%), 55 - 64 (11.5\%), 65+ (23.1\%), gender: female (54.4\%), male (43.4\%), other (2.2\%) and robot exposure frequency: Daily (19.8\%), Weekly (21.4\%), Monthly (14.8\%), Less Frequently (22.0\%), and Almost Never (22.0\%). Sample size sufficiency was determined by considering study criteria including broad exploratory scope, sparse cohort specificity, and limited established domain theory. Repeated iterative analysis showed saturation of results was reached, thus sample size was sufficient. Quantitative results regarding negative human-perceptions of robots for this same cohort can be viewed in the companion paper~\cite{Allen2024e1p1} and research project website$^1$.

\subsection{Topic Modelling Analysis} \label{sec:topic-analysis}
Participant comments were coded on three dimensions, question priming, sound properties commented on, and valence of comment. Question priming refers to whether the participant was aware of the context of research being sound or not (see section~\ref{sec:questions-and-priming}). From the 77 participants who experienced the sound condition and provided qualitative feedback, 61 (79.2\%) mentioned sound unprompted.

Auto-thematic coding and word cloud searches were used to identify 16 sound property topic codes based on prevalent re-occurring terms and their counterparts (e.g. both pitch high and low need to be coded for completeness despite one being mentioned more commonly). The 16 sound property codes appear in table~\ref{tab:sound-codes-with-examples} alongside example terms which (assuming the context of the full participant comment aligned) are likely to have been assigned to each code. When asked about sounds robots should produce, several people discussed voices they wanted robots to have, so these codes have been included to aid future researchers.

\begin{table}
    \centering
    \caption{Sound Property topic codes showing example terms which contextually would code to each topic.}
    \label{tab:sound-codes-with-examples}
    \setlength\tabcolsep{4pt}
    \begin{tabular}{|p{0.14\linewidth}|p{0.26\linewidth}|p{0.47\linewidth}|}
	\hline
	\textbf{Category}&\textbf{Sound Code}&\textbf{Example Terms}\\\hline
 
        \multirow{2}{0.14\linewidth}{Pitch (Frequency)} 
        & \textbf{Pitch High} & high frequency, high pitch, squeal\\\cline{2-3}
	& \textbf{Pitch Low} & low frequency, low pitch, deep sound\\\hline
 
        \multirow{4}{0.14\linewidth}{Timing} 
        & \textbf{Acute} & sharp, short, sudden, alert, rapid, temporal\\\cline{2-3}
	& \textbf{Constant} & always, consistent, steady, chronic, continuous, stable, incessant\\\cline{2-3}
        & \textbf{Rhythmic} & rhythm, repeating, pattern\\\cline{2-3}
        & \textbf{Phase In or Out} & phase, gradual, fade, recede \\\hline
        
        \multirow{6}{0.14\linewidth}{Type of Sound} 
        & \textbf{Broadspectrum} & white noise, hum, buzz\\\cline{2-3}
	& \textbf{Natural or Animal} & organic, animal, ocean, bird, cat, dog\\\cline{2-3}
        & \textbf{Machine} & artificial, electric, motor, whirring, mechanical, beep, clank\\\cline{2-3}
        & \textbf{Informational} & communicative, functional, purpose, explained, match movement\\\cline{2-3}
        & \textbf{Artificial Voice} & digital, ai-like, mech, bot\\\cline{2-3}
        & \textbf{Human Voice} & human-sounding, person's voice\\\hline
        
        \multirow{3}{0.14\linewidth}{Volume (Intensity)} 
        & \textbf{Loud} & loud, hurt ears, high decibel, intense\\\cline{2-3}
	& \textbf{Audible} & want to hear, need sound, proxemic, not muted, ambient, noticeable\\\cline{2-3}
        & \textbf{Quiet} & no sound, silent, completely quiet, little as possible, soft\\\hline
        
        \multicolumn{2}{|c|}{\textbf{Subjective Qualities}}& Positive: energetic, meditative, calm, gentle, peaceful, friendly, rustling, comforting, reassuring, familiar, background noise, cute\newline
        Negative: unpleasant, distracting, obtrusive, frantic, creepy, confusing, annoying\\\hline
    \end{tabular}
\end{table}

\begin{figure}[t]
    \centering
    \includegraphics[width=0.6\linewidth]{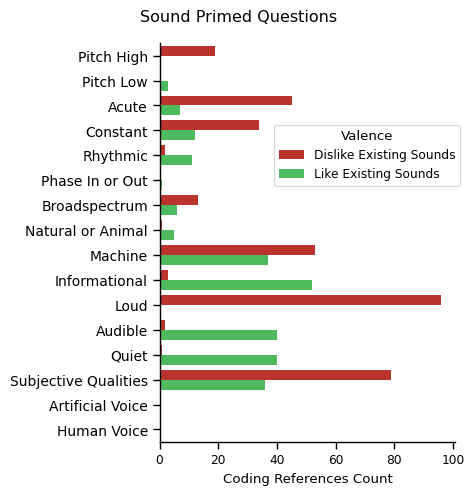}\hfill
    \caption{Sound primed responses - properties of existing consequential sounds which participants `like' and `dislike'.}
    \label{fig:code-freq-primed-existing}
\end{figure}

\begin{figure}[t]
    \centering
    \includegraphics[width=0.65\linewidth]{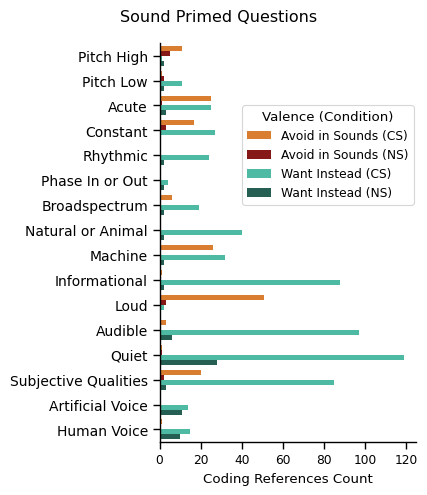}\hfill
    \caption{Sound primed responses - preferences for robot sounds, properties to `avoid' and what people `want instead', separated by sound (CS) versus no-sound (NS) participant condition.}
    \label{fig:code-freq-primed-preferred}
\end{figure}

\begin{figure}[t]
    \centering
    \includegraphics[width=0.6\linewidth]{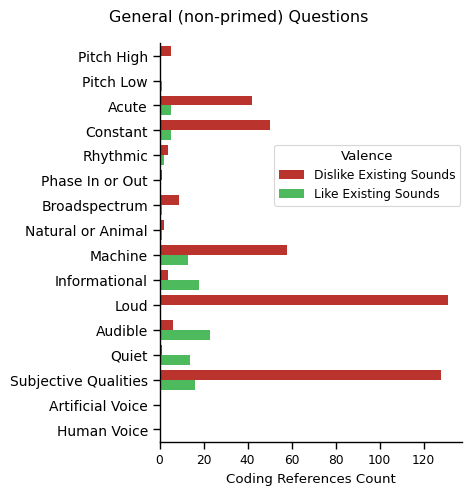}\hfill
    \caption{General (non-primed) responses - properties of existing consequential sounds which participants `like' and `dislike'.}
    \label{fig:code-freq-general-existing}
\end{figure}

\begin{figure}[t]
    \centering
    \includegraphics[width=0.66\linewidth]{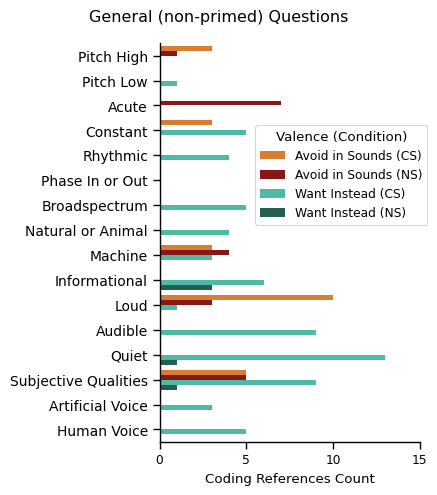}\hfill
    \caption{General (non-primed) responses - preferences for robot sounds, properties to `avoid' and what people `want instead', separated by participant sound condition (CS) versus (NS).}
    \label{fig:code-freq-general-preferred}
\end{figure}

In addition to sound property and question priming, comments were coded with one of 4 valences; `Like Existing Sounds' versus `Dislike Existing Sounds' (which considered the presented consequential sounds) and `Want Instead' versus `Avoid in Sounds' (which considered how participants would prefer robots to sound). After coding by the three dimensions, each coded reference was counted, with the results presented in figs.~\ref{fig:code-freq-primed-existing}, \ref{fig:code-freq-primed-preferred}, \ref{fig:code-freq-general-existing} and \ref{fig:code-freq-general-preferred}. Table~\ref{tab:participant-num-per-sound-code} shows how many individual participants provided qualitative responses related to each sound property code. Many participants' responses contained multiple references to the same sound property suggesting they had strong sentiments towards it, leading to longer bars for these sound properties in figs.~\ref{fig:code-freq-primed-existing}-\ref{fig:code-freq-general-preferred}. Interestingly, when unprompted, very few participants shared what sounds they liked or preferences for new sounds, but plenty shared dislikes, which were mostly loudness or subjective qualities. Figs.~\ref{fig:code-freq-primed-preferred} and \ref{fig:code-freq-general-preferred} contain breakdowns for preferred sounds by participant condition of sound (CS) versus no-sound (NS).

\begin{table}
    \centering
    \caption{Individual participants with qualitative responses related to each sound property showing number of participants with one or more references and \% of 182 participants.}
    \label{tab:participant-num-per-sound-code}
    \begin{tabular}{|p{0.28\linewidth}|p{0.22\linewidth}|p{0.21\linewidth}|}
		\hline
		\textbf{Sound Property} & \textbf{No. Participants} & \textbf{\% Participants} \\\hline
		Pitch High & 31 & 17.0\%\\\hline
		Pitch Low & 26 & 14.3\%\\\hline
		Acute & 75 & 41.2\%\\\hline
		Constant & 98 & 53.8\%\\\hline
		Rhythmic & 62 & 34.1\%\\\hline
		Phase In or Out & 27 & 14.8\%\\\hline
		Broadspectrum & 62 & 34.1\%\\\hline
		Natural or Animal & 63 & 34.6\%\\\hline
		Machine & 91 & 50.0\%\\\hline
		Informational & 80 & 44.0\%\\\hline
		Loud & 82 & 45.1\%\\\hline
		Audible & 76 & 41.8\%\\\hline
		Quiet & 97 & 53.3\%\\\hline
		Subjective Qualities & 103 & 56.6\%\\\hline
		Artificial Voice & 64 & 35.2\%\\\hline
		Human Voice & 76 & 41.8\%\\\hline
    \end{tabular}
\end{table}

\subsection{Sound Property Code Insights}
Valence perception of the 16 sound property codes contains many insights as to what properties are particularly liked (or wanted) versus disliked (or should be avoided).

\subsubsection{Pitch (Frequency)}
\textbf{Pitch High} was received poorly by most participants who commented, ``high pitch of the spinning rotors very strongly conveys danger'', with requests to ``keep the same sound but a little less high pitched''. Consistently with this, participants mentioning \textbf{Pitch Low} commented ``the sounds are lower pitched and less grating to my ear'' and suggested ``altering the sound to a lower frequency''.

\subsubsection{Timing}
Many participants reported disliking specific \textbf{Acute} sounds as ``the jolting sounds are agitating''. There were lots of requests for short sounds such as ``beeps'' and some observations on problems with mis-matched acute sounds, ``although the gestures convey meaning, the clunky noises distract from that". \textbf{Constant} sound received mixed positive comments ``background noise is nice'' and negative comments such as being glad that the ``sound wasn't constant''. Several comments suggested prolonged constant sound could be an issue, with some sounds being worse than others, ``constant electric motor noise wasn't so bad (e.g. the Jackal), but those combined with propeller sounds became a bit much if it were to be something I'm around for an extended period of time''. \textbf{Rhythmic} sound was primarily received well ``I like the thumping sound the feet make with the ground. It's rhythmic and energetic'', with suggestions for using rhythm to enhance sounds e.g. ``beeps indicating the speed at which its moving. Beeps get more frequent as the robot moves faster''. Whilst infrequently mentioned, \textbf{Phase In or Out} was detailed by several participants as a solution to consequential sounds e.g. ``if sounds of the motors are unavoidable, then it would be best if they fade in at the start of each movement and fade out at the end to make it less annoying.''

\subsubsection{Type of Sound}
\textbf{Broadspectrum} sounds showed more negative than positive responses, which is unexpected when compared to existing literature~\cite{Fastl2007a,Cha2018}. It was common to dislike specific broadspectrum sounds ``remove the white noise'', but request different broadspectrum sounds ``a low humming noise so you know they are on/moving, something like a heater/fan noise''. Some participants offered comparisons between two broadspectrum sounds ``my preference would be cat-purring-sound rather than the white-noise robot sound''. Many participants expressed wanting \textbf{Natural or Animal} sounds for robots, ``like the waves at the beach or something like rain would be nice''. Several comments indicated wanting more organic sounds ``similar to cats and dogs that don't make that much noise, maybe a quiet `padding' noise as they walk'', with one person remarking the ``closer a robot is mimicking or replacing interaction with a human or an animal, then we should only expect to hear sounds similar to a human or animal''. \textbf{Machine} sounds had mixed positive ``mechanical sounds of its movements is somewhat reassuring. Much like a mechanical keyboard'' and negative responses ``Softer, less machine-like sounds''. People tended to dislike the machine-like qualities, but appreciate what these sounds could tell you about the robot.

A significant number of participants reported wanting the sound to be \textbf{Informational} and contain details on the state of the robot. Examples included: ``I like that robots make sound. It helps to know what they are doing besides just the physical motion", wanting a sense of proxemics e.g. ``low sound of waves in the ocean (while it is moving) to calmly signal that the robot is around'' and cues for the robot's function by ``hav(ing) more functional sounds to reveal something about their function or intention (e.g. trucks making beeps when they are backing)''. Many participants gave unprompted opinions on wanting robots to have voices, either artificial or human. \textbf{Artificial Voice}'s were suggested such as a ``cute robo voice'' or to be able to ``respond with voice. But I would prefer that they would be distinguishable as a robot". A \textbf{Human Voice} was requested by several people, with suggestions for a ``voice quality of whatever language or accent it was speaking in'' and one participant being concerned that ``the lack of a humanlike voice made it seem like Pepper was being held hostage or perhaps was being suffocated''. Many of the participants in the no-sound condition requested voices, and may have made assumptions about what sounds robots do/can make due to not having a clear concept of consequential sounds.

\subsubsection{Volume (Intensity)}
As expected, most participants expressed negative views towards \textbf{Loud} sounds, often specifically mentioning `loud noise'. Several participants shared that loud sounds made them disinterested in robots, ``(the robot) is too loud that makes it much less desirable and would make me much less likely to want to interact with or use that robot", with one participant appealing ``I like robots and what the future will bring. But please make it with less noise''. Whilst not wanting loud robots, many participants wanted robots to be \textbf{Audible}, tying this to the desire for informative sounds e.g. to hear where a robot was, or what it was doing so they could ``sense it's completed its expected movement without having to look at it''. \textbf{Quiet} was also used in many responses, sometimes as simply as ``more quiet'' or ``less sound whilst operating, the better'', and a few people ``would prefer not to hear them at all''. Some participants provided insight into potential side effects of a quiet robot saying the ``quieter they are the easier it feels to trust them.  Quieter robots feel less intrusive, more helpful somehow''.

\subsubsection{Subjective Sound Qualities} 
Many participants included positive and/or negative subjective feedback that couldn't be objectively paired to a sound property, for example ``sounds that don't interfere with peace and/or distract from what you are doing''. A list of terms commonly mentioned by participants can be seen in table~\ref{tab:sound-codes-with-examples}.

\subsection{Linking Sound Features and Qualitative Responses} \label{sec:linking-to-spectrogram}
\begin{figure}[t]
    \centering
    \includegraphics[width=1\linewidth]{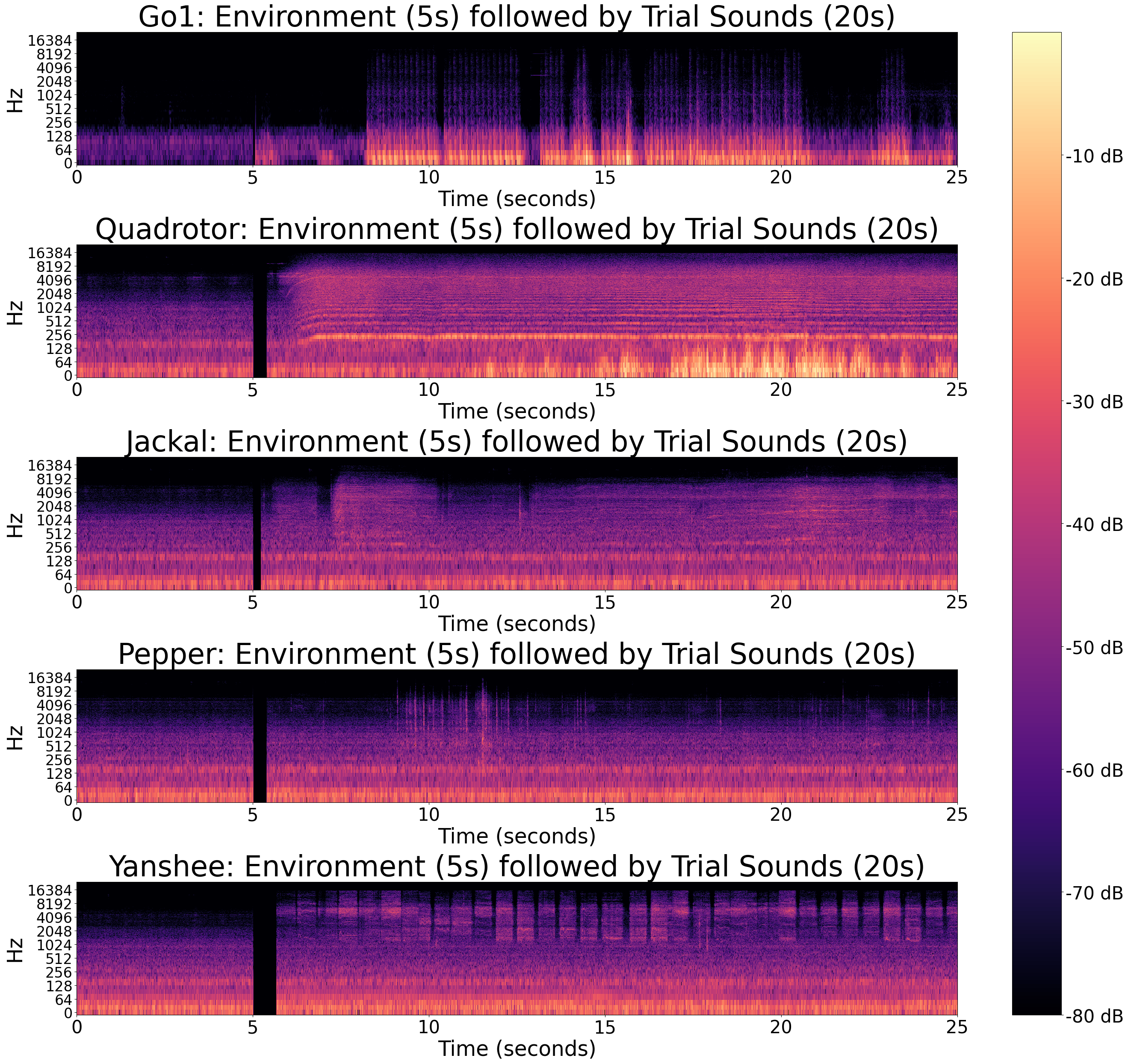}\hfill
    \caption{Spectrogram of experiment sound stimulus - the first 5 seconds show environmental sounds only (baseline), followed by a black no-sound separator, with the remaining 19-20 seconds containing the stimulus heard by participants i.e. robot consequential sounds + environment sounds.}
    \label{fig:spectrogram}
\end{figure}

Spectral features of robot consequential sounds can be observed in the robot videos$^1$ and via sound characterisation spectrograms of the robot video sound stimuli (provided in fig.~\ref{fig:spectrogram}). Sound intensity between each spectrogram has been scaled against the same peak intensity, thus allowing for between robot intensity (loudness) comparison. The spectrograms display changes in frequencies (y axis) of sounds produced over time (x axis). The intensity (amplitude) of each frequency is displayed as a colour relative to the maximum intensity (bright yellow-white), with less intense (perceptually quieter) sounds being pink or purple, and the least prevalent sound frequencies appearing black. The spectrograms show some commonalities across the robot sounds such as more power in the higher frequency ranges whilst the robot is moving.

When considering features disliked by participants, acute spikes into higher pitched frequencies (vertical lines) can be clearly seen across all spectrograms due to each robot's movements, and are particularly prevalent in the spectrograms of Yanshee, Go1 and Pepper. The contrast in intensity between the environment only sounds (first 5 seconds) and the increased intensity during the 20 seconds containing robot consequential sounds, clearly conveys the most disliked property of being loud. The sounds were likely perceived as loud relative to the environmental sounds, and possibly additionally due to subjective properties such as pleasantness or what information the noise was contributing. Participants reported liking audible sounds which contained information on the robot, which can clearly be observed in the spectrograms, for example short stop-start bursts from direction changes of the Jackal's wheels. Many of the robot consequential sounds follow regular spectral patterns as a limb moves or wheel spins, and this rhythmic sound was reported as liked by many participants. Some low pitch sounds (not commonly mentioned) can be observed for the Quadrotor and Go1 robots. All robots are clearly generating various broadspectrum sounds, with this property receiving mixed reviews from participants. The Quadrotor has distinct harmonic horizontal bands of buzzing from its rotors, which likely overpowered the rest of the sound spectrum.

\section{Discussion} \label{sec:discussion}
The results presented address the two research questions (see section~\ref{sec:research-questions}) providing insights into what properties of consequential sounds are liked and disliked, and what sounds people would prefer robots to make instead. Supporting prior research showing that sound is an important sensory channel~\cite{Jouaiti2019}, 79\% percent of participants responding to general (un-primed) questions referred to sounds of robots. When un-primed, participants commented on elements such as sounds being too loud or too high pitched, resulting in primarily negatively valenced comments. On the other hand, sound primed questions uncovered both positive and negative perceptions, as well as preferences for alternatives.

Considering volume (intensity) of sound, whilst quieter (not loud) sounds were preferred, many participants specifically requested audible sounds over completely silent robots. This desire for an audible proxemic robot sound was heavily entwined with wanting robots to produce informative sounds, providing feedback on position, purpose, or state of robot, and to aid in predictability of robot actions and trajectory. Preferences for timing of sounds showed a desire for rhythmic sounds, irrespective of whether those sounds were additionally mostly continuous, or acute (such as beeps). Acute sounds are to be avoided (except for when intentionally alerting a human of something), with some participants disliking completely continuous sounds without any changes over time. Whilst mentioned infrequently, phasing in and out of sounds had only positive perceptions. Combining these insights suggests that a rhythmic sound designed to cover gaps between acute sounds may be effective.

In terms of type of sound, many people had a preference for natural or animal sounds, with suggestions including cat purrs, dog barks and padding, ocean waves, and natural wind sounds. Machine sounds received mixed responses with some negatively associating them with loudness, distraction or fear, and other people liking more subtle machine sounds as long as they made sense for the type of robot. Informational sounds were considered positive including wanting non-distracting, background, ambient sounds for proxemics (similar to hearing a human breathing or rustling nearby) and acute informative beeps or tones. Another common request was for the robot to have a voice, with mixed opinions on if this should be human or artificial. Opinions on broadspectrum sounds varied dramatically, likely due to this profile covering quite a large range of sounds. There was some consistency if looking at specific broadspectrum sounds, such as a general like of natural cat purring and waves, and a general dislike of standard white noise often associated with static, whir, or buzz descriptions. Some participants compared two broadspectrum sounds as disliking one and wanting another, suggesting more research is needed on which broadspectrum sounds may be generally well received. Combining this with how pitch was perceived (low pitch generally well accepted but avoiding high pitch sounds), suggests there may be potential for more down-shifted (primarily lower frequency) broadband sounds. More detailed and specific sound preference responses were obtained from participants in the sound condition, suggesting the importance for robot sound design of exposing people to the current-state before asking for sound preferences.

Many subjective opinions on sounds and how the sounds effected them were shared by participants. Many expressed liking familiar sounds, however what was familiar for each person varied from kids toys, to robots they've used, or media depictions of robots. Subjective elements require further research, and will need to be considered when designing or improving robot sounds, especially if looking to personalise robot sounds to individual people. Whilst consequential sounds must exist, they may be modified to alter how people perceive them. The results from this paper provide actionable insights for sound design for robots. Our findings inform what sounds to alter (e.g. acute sounds) versus keep (e.g. informative), with preferred added (or removed) sounds informing choice of sound augmentation method, e.g. selecting robot components which generate more pleasant consequential sounds, utilising dampening or noise cancelling techniques (to reduce loudness), or adding sounds to disguise unwanted attributes.

Research presented in this paper is consistent with related literature which suggests that consequential sounds have a negative impact on HRI and perception of robots~\cite{Frid2018,Allen2024e1p1,Jouaiti2019,Trovato2018}. The findings agree with psychoacoustics theory relating to proxemics (wanting audible and informational  sounds)~\cite{Jouaiti2019,Trovato2018}, known psychological responses to different sound characteristics (avoiding high pitch, acute and loud noises)\cite{Cha2018, Jouaiti2019}, and the role of sound in multi-modal perception (preference for rhythmic and informational sounds and wanting voices for certain robots)\cite{DeGelder2000,Vroomen2000}. One interesting difference was that broadspectrum sounds had more negative than positive mentions, which is in contrast to some previous literature~\cite{Fastl2007a,Cha2018}, but can likely be attributed to the large range of sounds that can be classified as broadspectrum e.g. whirring motors are quite different to ocean waves.

There are several limitations of the results presented in this paper. Firstly, the study was conducted online, where audio and context cues are not as salient as in a physical room. Whilst prior HRI research suggests that it is likely that real-world results would show an even stronger negative correlation~\cite{DePaivaVianna2015,Jouaiti2019,Izui2020}, we cannot be certain if expressed opinions would be the same in a real world environment. Secondly, interaction context can be important to how robots are perceived ~\cite{Schute2007,DePaivaVianna2015,Allen2023}, so whilst some participants provided details (at home, whilst working etc), we cannot be sure what context participants imagined themselves being around the robot in. Findings could also change with purpose of interaction e.g. direct interaction versus colocation, or intent to alert the human. Thirdly, this paper mostly considered commonalities of perceptions across whole population and contexts. Whilst this provides a great starting point for generic improvements, this could be refined through further research on contexts and properties of individuals that could have an effect on variation in sound perception, preferences and interpretation. For example cultures with pitch sensitive languages might be more affected by high or low pitched robot sounds\cite{Wong2012}. Understanding individual differences would facilitate future customisation to specific stakeholders.

\section{Conclusions and Future Work}
How people feel around robots i.e. their human-centric perceptions of robots, is a critical concern if people are to co-habitate shared spaces with robots. This paper examined data from 182 participants sharing their qualitative views on how they perceive robots and their sounds. Participants watched videos of 5 robots and gave feedback on their perceptions of the robots, initially uninformed about the premise of the research being sound, and later with specific sound-primed questions. Topic analysis identified 16 key sound properties that people shared views on, and mapped feedback on these properties to 4 valences, like versus dislike current sounds, and what sounds they would prefer to have or avoid instead. Aside from the expected ubiquitous opinion of wanting less loud or high-pitched sounds, participants showed a clear preference for wanting robots to be audible as they contained valuable, informative, proxemic details such as location of the robot, and its state or processes. Further insights included a preference for rhythmic sounds and avoiding acute sounds. Machine-like sounds were generally perceived poorly, with many requesting more natural or animal sounds including cats, dogs, wind and ocean. Many people additionally expressed wanting robots to have a voice, especially if the robot resembled an animal or human. 

The insights gained from this research can assist robot designers to improve consequential sounds, and hence human-perception of robots. As expressed by one participant, ``In other words, consequential sounds are actually necessary -as long as it's fine tuned to the right pitch, tone, rhythm to make it more acceptable to the human ear. I'd suggest more research in the latter.'' Potential areas for future research include examining the combined effects on human-perception of robot consequential sounds with different ambient sounds, and exploratory analysis on correlations between human perceptions, sound properties and specific groups of people and contexts e.g. introverts vs extraverts or working versus a social gathering. Building on this work, research should be done on how commonly disliked features can be reduced, and preferred sound properties added, i.e. new methods for targeting and improving consequential sounds (for example sound augmentation), leading to improved human-perception towards robots.

\IEEEtriggeratref{20}
\bibliographystyle{IEEEtran}
\bibliography{arxiv}

\end{document}